\documentclass[acmtog]{acmart}

\usepackage{booktabs} 

\usepackage[ruled]{algorithm2e} 

\SetAlFnt{\small}
\SetAlCapFnt{\small}
\SetAlCapNameFnt{\small}
\SetAlCapHSkip{0pt}

\acmJournal{TOG}

\begin{document}
\title{Aesthetics Driven Autonomous Time-Lapse Photography Generation by Virtual and Real Robots}


\author{Xiaobo Gao}
\affiliation{%
  \institution{Beihang University}
  \city{Beijing}
  \country{China}}
\email{gaoxiaobo@buaa.edu.cn}

\author{Qi Kuang}
\orcid{0000-0002-1675-1197}
\affiliation{%
  \institution{Beihang University}
  \city{Beijing}
  \country{China}}
\email{kuangqi@buaa.edu.cn}

\author{Xin Jin}
\orcid{0000-0003-3873-1653}
\affiliation{%
  \institution{Beijing Electronic Science and Technology Institute}
  \city{Beijing}
  \country{China}
}
\email{jinxinbesti@foxmail.com}

\author{Bin Zhou}
\orcid{0000-0003-2091-1090}
\affiliation{%
  \institution{Beihang University}
  \city{Beijing}
  \country{China}}
\email{zhoubin@buaa.edu.cn}

\author{Boyan Dong}
\affiliation{%
  \institution{Beijing Electronic Science and Technology Institute}
  \city{Beijing}
  \country{China}
}
\email{513935176@qq.com}

\author{Xunyu Wang}
\affiliation{%
  \institution{Beijing Electronic Science and Technology Institute}
  \city{Beijing}
  \country{China}
}
\email{chevalierwang@outlook.com}

\begin{abstract}
Time-lapse photography is employed in movies and promotional films because it can reflect the passage of time in a short time and  strengthen the visual attraction. However, since it takes a long time and requires the stable shooting, it is a great challenge for the photographer.
  In this article, we propose a time-lapse photography system with virtual and real robots. To help users shoot time-lapse videos efficiently, we first parameterize the time-lapse photography and propose a parameter optimization method. For different parameters, different aesthetic models, including image and video aesthetic quality assessment networks, are used to generate optimal parameters. Then we propose a time-lapse photography interface to facilitate users to view and adjust parameters and use virtual robots to conduct virtual photography in a three-dimensional scene. The system can also export the parameters and provide them to real robots so that the time-lapse videos can be filmed in the real world.
  In addition, we propose a time-lapse photography aesthetic assessment method that can automatically evaluate the aesthetic quality of time-lapse video.
  The experimental results show that our method can efficiently obtain the time-lapse videos. We also conduct a user study. The results show that our system has the similar effect as professional photographers and is more efficient.
\end{abstract}

%
%


%
%


\maketitle

\section{Introduction}
With the popularity of smartphones and cameras, more and more people are demanding the high-quality photography. Therefore, many research works focus on image and video shooting to provide useful guidance \cite{rawat2016clicksmart, rawat2019photography, lou2021aesthetic}.
Time-lapse photography is one of the special photographic techniques because of its ability to present the low-speed changes in a scene with the high playback speed \cite{bennett2007computational}.
However, photographers have to spend hours or even days shooting to get the final time-lapse video clips.
And during the shooting, the camera needs to be stationary or move slowly and steadily. The photographers need to adjust the camera pose frequently, which is time-consuming and laborious for the photographers.
As a result, robots are more suitable for this kind of time-consuming work that requires stable movement because of the stable pan-tilt and precise control system.

However, it is difficult to make the robots shoot autonomously. Successful photography usually requires the photographers to have rich shooting experience and strong aesthetic ability. How to determine a series of parameters and let the robots shoot automatically or assist the users in shooting is a difficult point.
Computational aesthetics is the use of machines to achieve a computable human aesthetic process \cite{neumann2005defining}, so that computational methods can make applicable aesthetic decisions in a similar way to humans.
Therefore, the system with the combination of robots and computational aesthetics makes automatic time-lapse photography possible.

This paper proposes an automatic time-lapse photography framework, which conducts the virtual and real-world time-lapse photography with robots by calculating the time-lapse photography parameters of the three-dimensional scene.
We propose a parameter optimization method to obtain the optimal shooting parameters. The innovation is to parameterize the complex process of video generation of time-lapse photography. We classify the parameters into image aesthetics-related, video aesthetics-related, and time-lapse aesthetics-related according to the characteristics of time-lapse photography, and propose parameter optimization methods for these three types of parameters. According to the characteristics of different parameters, we design and employ different aesthetic models for parameter optimization.

Meanwhile, we propose a time-lapse photography interface for users to view or adjust the parameters generated automatically by our method and use virtual robots to take virtual photography in three-dimensional scenes.
Users can preview the time-lapse photography videos, and adjust the parameters according to their own preferences.
The system also supports the export of parameters and provides them to real robots to enable robots to move and control the camera in the real world, shooting time-lapse videos with specific time parameters.In addition, we propose a time-lapse photography aesthetic evaluation method that can automatically evaluate the aesthetic quality of time-lapse video. We apply it to validate our parameter optimization method.

The main contributions of this work can be summarized as follows:
\begin{itemize}
\item a time-lapse photography framework that can automatically and efficiently shoot time-lapse videos.
\item a method for time-lapse photography parameter optimization using different aesthetic quality assessment models.
\item a time-lapse video assessment method to evaluate the aesthetic quality of time-lapse video.
\item a time-lapse photography interface for users to conduct the virtual or real-world time-lapse photography.
\end{itemize}

\section{Related Work}

Existing work on time-lapse photography has focused more on generating optimised time-lapse videos from existing images or videos, i.e. the post-processing of the data. There is no research on how to go about capturing time-lapse video or the acquisition of the data.

\begin{figure*}
\begin{center}
  \includegraphics[width=\linewidth]{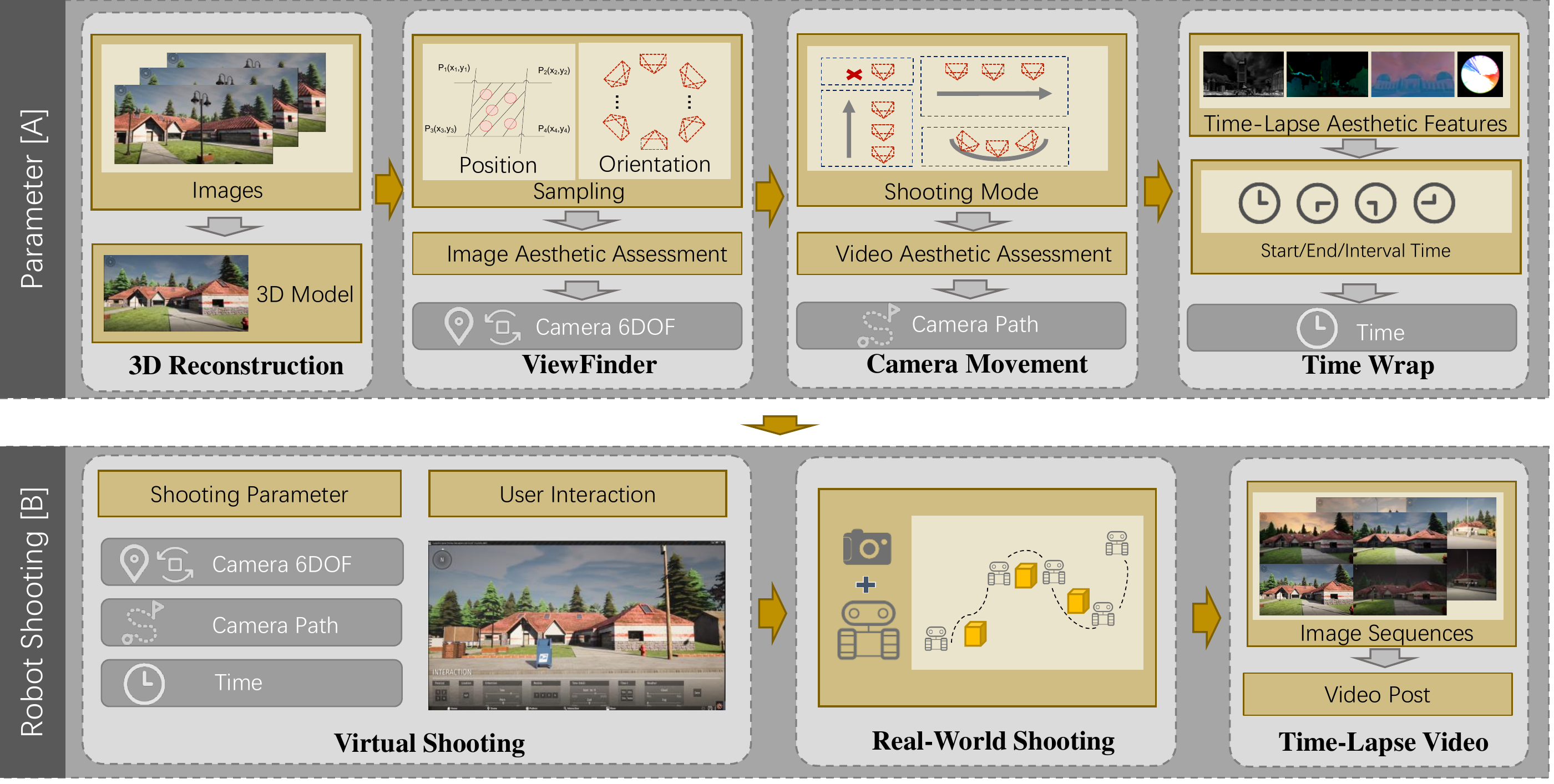}\\
\end{center}
  \caption{Illustration of the proposed framework. The 3D scene is reconstructed with the images. The shooting parameters are then determined by analyzing the 3D scene in the simulation environment. The interactive interface enables users to modify or view parameters to complete virtual photography. The system can also provide parameters for real robots to complete real-world time-lapse photography.}
\label{fig2}
\end{figure*}

\section{Method}

To obtain the satisfactory time-lapse videos, careful preparations are needed for photographers in advance, including scenery selection, shooting plan, etc. Moreover, time-lapse photography requires photographers to take a large number of images over a long period of time, which requires a great deal of patience and carefulness.
Therefore, in order to generate the optimal shooting plan so that the robot can automatically generate the time-lapse videos or assist the user in photography, our goal is to design a time-lapse shooting system that can find a series of optimal parameters resulting in the high quality of the final video. So we define the overall quality function $Q$ as follows:
\begin{equation}
\label{E1}
Q = Q_{I}(\mathcal{P_{I}}) + Q_{V}(\mathcal{P_{V}}) + Q_{T}(\mathcal{P_{T}}),
\end{equation}
where $Q_{I}$, $Q_{V}$, $Q_{T}$ are the quality functions for image, video and time-lapse video respectively. $\mathcal{P_{I}}$, $\mathcal{P_{V}}$, $\mathcal{P_{T}}$ are the shooting parameters.

Secondly, different users may have different preferences when shooting time-lapse videos. The system should therefore support parameter personalization to suit different users.
Moreover, in case of automatic robot shooting, it also involves a series of parameters of the robot.

In order to solve the above difficulties, we propose an automatic time-lapse photography framework, as shown in Figure~\ref{fig2}.

\textbf{3D Reconstruction.}
When shooting a scene, experienced photographers will usually investigate the scene in advance, which takes a lot of time to determine the location and plan the shot. The transformation of light and shadow is a very important factor in time-lapse photography, so the shooting time is also important.
In order to capture the desired light and shadow, some photographers need to rehearse the plan in advance to ensure that they can efficiently conduct the photography at a specific time during the actual shooting.
Therefore, adequate preparation is very important for photography, especially time-lapse photography.

3D reconstruction allows us to quickly understand an unknown scene and react accordingly. The analysis of a 3D model in a simulation environment is an effective way to fleetly determine the parameters for time-lapse photography based on the shooting scene.

First we reconstruct the shooting scene. For large-scale scenes, image-based 3D reconstruction is a low-cost and efficient reconstruction method. The images can be collected on the Internet or captured by drone or robot.
We then use structure-from-motion, an image-based 3D reconstruction method, to obtain a textured and realistic 3D scene. In order to obtain reachable shooting points, we limit the shooting height and generate reachable areas.

The image-based reconstructed 3D scene is static, but dynamic objects are important in time-lapse photography. So we add dynamic elements to the static 3D scene, mainly including people, traffic, weather and lighting.
Accessible regions on the ground are chosen to simulate the flow of people and vehicles to ensure that the dynamic elements are similar to those in the real world.

\textbf{Shooting Parameter.}
For time-lapse photography we first determine the relevant parameters, including shooting location $L$, shooting angle $A$, shooting path $P$, time range $[S_{t}, E_{t}]$, frame extraction interval $\Delta t$.
As shown in Equation~\ref{E1}, the overall video quality is mainly related to the image quality, ordinary video quality, and time-lapse video quality. Therefore, we classify the relevant parameters into three categories as well.

The first category is the viewfinder parameters $\mathcal{P_{I}}=\{L,A\}$, which are related to the image quality, including the shooting location $L$ and shooting angle $A$.
The second category is the camera movement parameters $\mathcal{P_{V}}=\{P\}$ that are related to the ordinary video quality since the camera motion of time-lapse photography is very similar to the camera motion of ordinary photography.
The last category is the time-warp parameters $\mathcal{P_{T}}=\{[S_{t}, E_{t}],\Delta t\}$, including time range $[S_{t}, E_{t}]$ and time interval $\Delta t$, which are the characteristic parameters of time-lapse video to reflect the time change.

After obtaining the 3D model we can analyze the geometry and texture of the 3D model, as well as the image or video rendered under different lighting conditions, in order to determine the optimal set of shooting parameters.

\textbf{Video Post. }
After determining the relevant parameters, we can determine the initial position of the camera according to the shooting position and orientation, then determine the camera movement according to the shooting path, and finally make the camera take a set of photos according to the time range and interval to get the final time-lapse video.

The flicker might occur when the final time-lapse videos are created \cite{martin2015time}.
Essentially instead of the camera’s settings between shots remaining perfectly constant or purposefully changing in a slight and gradual way, large unintended exposure jumps occur in a few of the frames which create images that look out of place when compiled with other images.
One of the ways that reduce the likelihood of flicker to occur is shooting in manual mode \cite{chylinski2012time}. However, since the light is changing in our scenes and manual mode might quickly over or under expose the images as the scene changes, we still use auto exposure mode, which makes it necessary to post-process the captured images to reduce flicker.
We smooth out the exposure in the time-lapse video with histogram equalization \cite{pizer1987adaptive} to create a good visual effect.
Finally we generate the time-lapse video with the processed image sequences.

It can be seen that the results using the parameters selected by the system are better than those selected by the amateurs. In the first two scenarios, the time-lapse videos generated by our method are of higher quality than those captured manually. The video quality captured by the professional photographers in the latter two scenes is the highest. The overall results show that our approach is better than the amateurs and comparable to the professional photographers. Therefore, it can be considered that our system can help amateurs to get more professional time-lapse videos. And our system can automatically generate high-quality time-lapse videos, which also saves photographers a lot of time.

\section{Results}
In this section we present two applications based on our system, namely virtual and real-world shooting.
\subsection{Virtual Shooting}

The system supports the import of 3D scenes, which can be obtained either automatically and quickly through 3D reconstruction or through manual modeling (which is time-consuming but more detailed). In the imported 3D scene, users can take virtual videos, which can be applied to film and television production, scene roaming, etc. The virtual photography can be classified into automatic photography and user-defined photography.

\subsubsection{Automatic Shooting}
Users can just import the appropriate 3D scenes and the system can automatically obtain the optimal parameters and generate video results for preview with our method.

\subsubsection{Customization Parameters}
\begin{figure}
\begin{center}
  \includegraphics[width=\linewidth ]{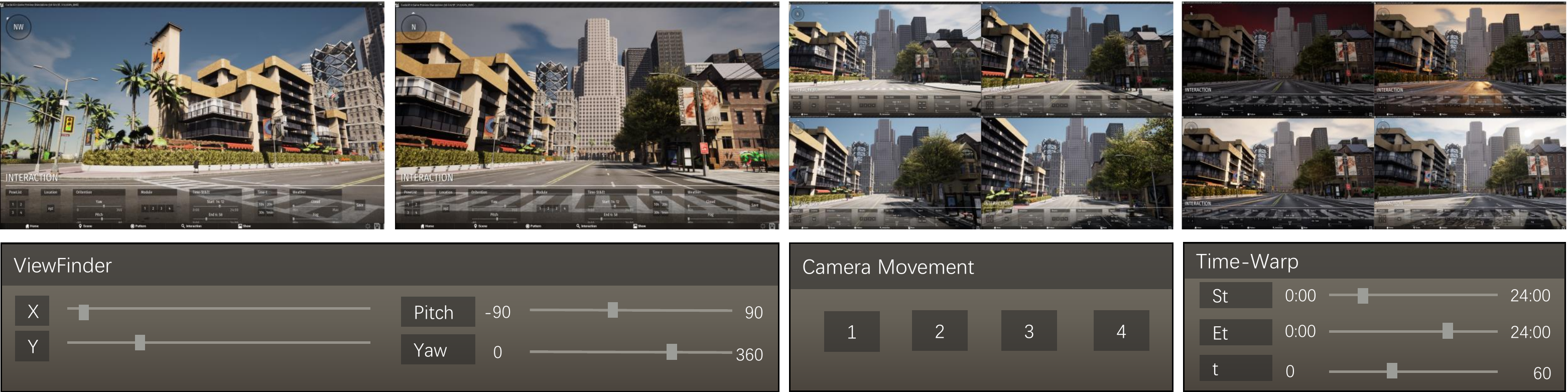}\\
\end{center}
  \caption{Virtual shooting. The first line is the visual interface, and the second line is the interface for parameter adjustment. From left to right: camera position, camera orientation, camera movement and time parameters.}
\label{fig7}
\end{figure}

In addition, the system supports user-defined parameters and modification of parameters so that the photography process can be done manually or semi-automatically.
As shown in the Figure ~\ref{fig7}, users can click and drag to complete the virtual scene selection, and then choose the appropriate mode of camera movement from the four shooting modes.
After that, users can set the time parameter to preview the lighting change instantly without waiting for a long time. The instant feedback from the interactive system makes it easier for users to set parameters.

\subsection{Real-World Shooting}

The system also supports outputting parameters and providing them to the robotics platform to shoot time-lapse videos in real environments.

\begin{figure}
\begin{center}
  \includegraphics[width=\linewidth]{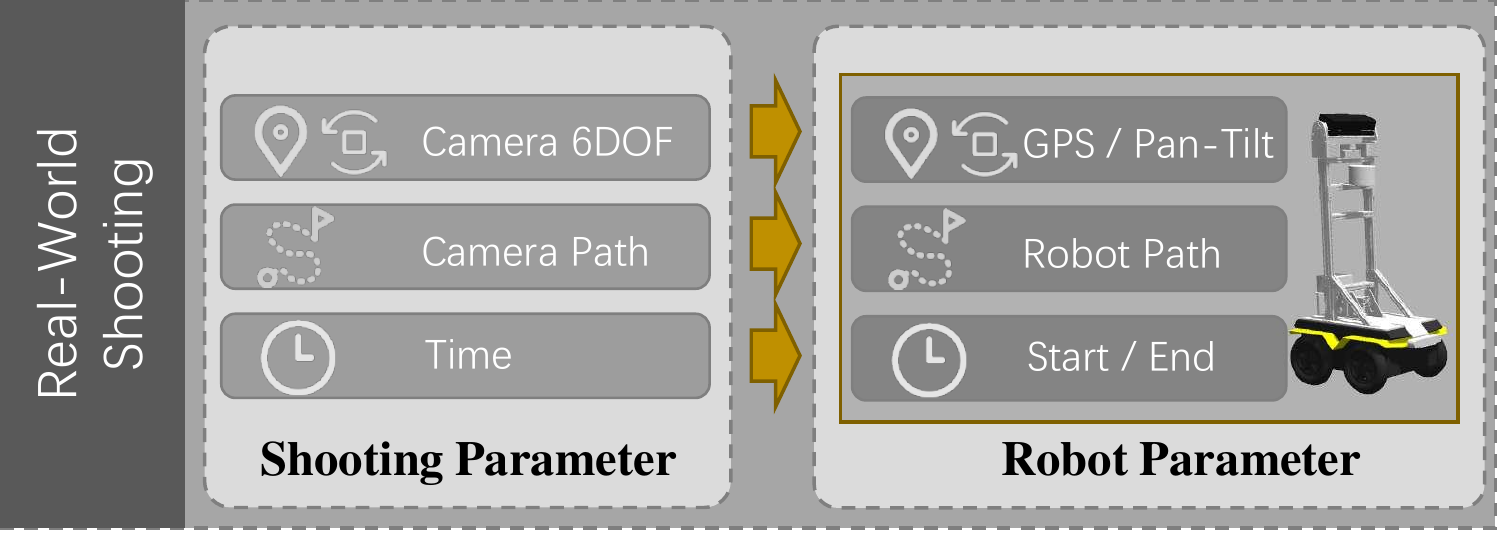}\\
\end{center}
  \caption{Real-World shooting. The parameters generated by the system need to be converted into the parameters of the robot.}
\label{fig8}
\end{figure}

As shown in the Figure ~\ref{fig8}, the parameters of the viewfinder are the camera position and orientation, which are converted into GPS coordinates and the pitch and yaw angles of the pan-tilt. The parameter of the camera motion is the motion path, which is converted into the motion trajectory of the robot. The time parameters are the time range and time interval, which we convert into the start and end time of the robot and the shooting interval. After the robot receives the relevant parameters, it will shoot on the spot according to the shooting plan.

\subsection{Evaluation}
\subsubsection{Self-Evaluation.}
The parameters associated with time-lapse photography are classified into three types of parameters: image, video and time-lapse aesthetics-related. In order to verify the validity of each type of parameter, we conduct the self-evaluation experiment.

We choose five simulated environments, each divided into four areas to be filmed, so that a total of 20 time-lapse videos are taken. The resulting videos are scored using our time-lapse video aesthetic quality assessment method.
We contrast our method of parameter selection with random parameter selection. To get the overall results for random parameters, we randomly select 10 sets of parameters and average them. We first test the results of random selection of all parameters, and then use our method to select parameters in the order of image, video, and time.

Table 1 shows the experimental results of the self-evaluation. It can be seen that our system progressively reaches the optimum in determining the parameters. For every type of parameter, our optimization method outperforms the random selection method.

\subsubsection{Comparison with Photographers.}
In order to verify the effectiveness of the overall system, we invited two professional photographers with rich experience and two novices who lack photography experience to manually set the parameters in the simulation scene and take photography.
We selected five different shooting areas. The user needs to select the appropriate position and orientation in the area to be shot, and determine the way of camera movement and the shooting time. Finally, the system is used for virtual photography to generate a time-lapse video.
At the same time, the system also automatically takes virtual photography in the scene with our optimized parameters.
The time-lapse photography aesthetic quality evaluation model is utilized to score the final time-lapse photography video, and the results are shown in the Table 2.

It can be seen that the results using the parameters selected by the system are better than those selected by the amateurs. The overall results show that our approach is better than the amateurs and comparable to the professional photographers. Therefore, it can be considered that our system can help amateurs to get more professional time-lapse videos. And our system can automatically generate high-quality time-lapse videos, which also saves photographers a lot of time.

\begin{table}
\small
\caption{Alation study. The check mark indicates that our method was used for parameter optimization. The unmarked term means random selection of parameters.}\label{table1}
\begin{center}
\begin{tabular}{ccccc}
  \toprule
         &   Image  &   Video  &   Time   &  Quality\\
  \midrule
  1      &          &          &          &  0.508\\
  2      &\checkmark&          &          &  0.522\\
  3      &\checkmark&\checkmark&          &  0.574\\
  4      &\checkmark&\checkmark&\checkmark&\textbf{0.635}\\
  \bottomrule
\end{tabular}
\end{center}
\end{table}

\begin{table}
\small
\caption{Comparison results with the shots of photographers.}\label{table2}
\begin{center}
\begin{tabular}{ccccccc}
  \toprule
        Scene  &   Amateur  &   Professional   &  Ours\\
  \midrule
        1      &  0.420     &  0.577     &\textbf{0.582} \\
        2      &  0.485     &  0.567     &\textbf{0.596} \\
        3      &  0.486     &\textbf{0.580}     & 0.537 \\
        4      &  0.479     &\textbf{0.547}     & 0.505 \\
  \midrule
        Avg.   &  0.468    &\textbf{0.568}    & 0.555 \\
  \bottomrule
\end{tabular}
\end{center}
\end{table}

\section{Summary}
In this paper, we propose an aesthetics driven framework for automatic time-lapse photography, which replaces photographers with virtual and real robots to complete the time-consuming work of time-lapse photography.
First, we parameterized the complex time-lapse photography process and proposed a method to optimize the parameters with the aesthetic models.
Then we proposed a specific time-lapse video aesthetic quality evaluation model to automatically evaluate the aesthetic quality of time-lapse photography.
In addition, we also proposed an automatic time-lapse photography system that supports users to manually, semi-automatically or automatically complete virtual photography. The system can also import parameters to real robots to complete real photography.
Experimental results and user studies show that our method can efficiently complete time-lapse photography automatically, and the effect is equivalent to that of professional photographers, which proves that our system can help inexperienced users to shoot high-quality time-lapse videos.

\bibliographystyle{ACM-Reference-Format}
\bibliography{sample-bibliography}


\begin{thebibliography}{8}


\ifx \showCODEN    \undefined \def \showCODEN     #1{\unskip}     \fi
\ifx \showDOI      \undefined \def \showDOI       #1{#1}\fi
\ifx \showISBNx    \undefined \def \showISBNx     #1{\unskip}     \fi
\ifx \showISBNxiii \undefined \def \showISBNxiii  #1{\unskip}     \fi
\ifx \showISSN     \undefined \def \showISSN      #1{\unskip}     \fi
\ifx \showLCCN     \undefined \def \showLCCN      #1{\unskip}     \fi
\ifx \shownote     \undefined \def \shownote      #1{#1}          \fi
\ifx \showarticletitle \undefined \def \showarticletitle #1{#1}   \fi
\ifx \showURL      \undefined \def \showURL       {\relax}        \fi
\providecommand\bibfield[2]{#2}
\providecommand\bibinfo[2]{#2}
\providecommand\natexlab[1]{#1}
\providecommand\showeprint[2][]{arXiv:#2}

\bibitem[Bennett and McMillan(2007)]%
        {bennett2007computational}
\bibfield{author}{\bibinfo{person}{Eric~P Bennett} {and}
  \bibinfo{person}{Leonard McMillan}.} \bibinfo{year}{2007}\natexlab{}.
\newblock \showarticletitle{Computational time-lapse video}.
\newblock In \bibinfo{booktitle}{\emph{ACM SIGGRAPH 2007 papers}}.
  \bibinfo{pages}{102--es}.
\newblock


\bibitem[Chylinski(2012)]%
        {chylinski2012time}
\bibfield{author}{\bibinfo{person}{Ryan Chylinski}.}
  \bibinfo{year}{2012}\natexlab{}.
\newblock \bibinfo{booktitle}{\emph{Time-lapse photography: A Complete
  Introduction to Shooting, Processing, and Rendering Time-lapse Movies with a
  DSLR Camera}}.
\newblock \bibinfo{publisher}{LearnTimelapse. com}.
\newblock


\bibitem[Lou et~al\mbox{.}(2021)]%
        {lou2021aesthetic}
\bibfield{author}{\bibinfo{person}{Hao Lou}, \bibinfo{person}{Heng Huang},
  \bibinfo{person}{Chaoen Xiao}, {and} \bibinfo{person}{Xin Jin}.}
  \bibinfo{year}{2021}\natexlab{}.
\newblock \showarticletitle{Aesthetic Evaluation and Guidance for Mobile
  Photography}. In \bibinfo{booktitle}{\emph{Proceedings of the 29th ACM
  International Conference on Multimedia}}. \bibinfo{pages}{2780--2782}.
\newblock


\bibitem[Martin-Brualla et~al\mbox{.}(2015)]%
        {martin2015time}
\bibfield{author}{\bibinfo{person}{Ricardo Martin-Brualla},
  \bibinfo{person}{David Gallup}, {and} \bibinfo{person}{Steven~M Seitz}.}
  \bibinfo{year}{2015}\natexlab{}.
\newblock \showarticletitle{Time-lapse mining from internet photos}.
\newblock \bibinfo{journal}{\emph{ACM Transactions on Graphics (TOG)}}
  \bibinfo{volume}{34}, \bibinfo{number}{4} (\bibinfo{year}{2015}),
  \bibinfo{pages}{1--8}.
\newblock


\bibitem[Neumann et~al\mbox{.}(2005)]%
        {neumann2005defining}
\bibfield{author}{\bibinfo{person}{L Neumann}, \bibinfo{person}{M Sbert},
  \bibinfo{person}{B Gooch}, \bibinfo{person}{W Purgathofer}, {et~al\mbox{.}}}
  \bibinfo{year}{2005}\natexlab{}.
\newblock \showarticletitle{Defining computational aesthetics}.
\newblock \bibinfo{journal}{\emph{Computational aesthetics in graphics,
  visualization and imaging}} (\bibinfo{year}{2005}), \bibinfo{pages}{13--18}.
\newblock


\bibitem[Pizer et~al\mbox{.}(1987)]%
        {pizer1987adaptive}
\bibfield{author}{\bibinfo{person}{Stephen~M Pizer}, \bibinfo{person}{E~Philip
  Amburn}, \bibinfo{person}{John~D Austin}, \bibinfo{person}{Robert Cromartie},
  \bibinfo{person}{Ari Geselowitz}, \bibinfo{person}{Trey Greer},
  \bibinfo{person}{Bart ter Haar~Romeny}, \bibinfo{person}{John~B Zimmerman},
  {and} \bibinfo{person}{Karel Zuiderveld}.} \bibinfo{year}{1987}\natexlab{}.
\newblock \showarticletitle{Adaptive histogram equalization and its
  variations}.
\newblock \bibinfo{journal}{\emph{Computer vision, graphics, and image
  processing}} \bibinfo{volume}{39}, \bibinfo{number}{3}
  (\bibinfo{year}{1987}), \bibinfo{pages}{355--368}.
\newblock


\bibitem[Rawat and Kankanhalli(2016)]%
        {rawat2016clicksmart}
\bibfield{author}{\bibinfo{person}{Yogesh~Singh Rawat} {and}
  \bibinfo{person}{Mohan~S Kankanhalli}.} \bibinfo{year}{2016}\natexlab{}.
\newblock \showarticletitle{ClickSmart: A context-aware viewpoint
  recommendation system for mobile photography}.
\newblock \bibinfo{journal}{\emph{IEEE Transactions on Circuits and Systems for
  Video Technology}} \bibinfo{volume}{27}, \bibinfo{number}{1}
  (\bibinfo{year}{2016}), \bibinfo{pages}{149--158}.
\newblock


\bibitem[Rawat et~al\mbox{.}(2019)]%
        {rawat2019photography}
\bibfield{author}{\bibinfo{person}{Yogesh~Singh Rawat},
  \bibinfo{person}{Mubarak Shah}, {and} \bibinfo{person}{Mohan~S Kankanhalli}.}
  \bibinfo{year}{2019}\natexlab{}.
\newblock \showarticletitle{Photography and Exploration of Tourist Locations
  Based on Optimal Foraging Theory}.
\newblock \bibinfo{journal}{\emph{IEEE Transactions on Circuits and Systems for
  Video Technology}} \bibinfo{volume}{30}, \bibinfo{number}{7}
  (\bibinfo{year}{2019}), \bibinfo{pages}{2276--2287}.
\newblock


\end{thebibliography}

\end{document}